\documentclass[runningheads,a4paper]{llncs}

\usepackage[utf8]{inputenc}

\let\llncssubparagraph\subparagraph
\let\subparagraph\paragraph
\usepackage[compact]{titlesec}
\let\subparagraph\llncssubparagraph

\usepackage{amssymb}
\usepackage{graphicx}
\usepackage{subfig}
\usepackage{booktabs}
\usepackage{multirow}
\usepackage{color}
\usepackage[misc]{ifsym}
\usepackage{bbding}
\usepackage[compact]{titlesec}

\usepackage{url}

\begin{document}

\mainmatter 

\title{Recurrent Neural Networks\\for Online Video Popularity Prediction}
\titlerunning{Recurrent Neural Networks for Online Video Popularity Prediction}

\author{Tomasz Trzci\'{n}ski\inst{1, \, 2}, Paweł Andruszkiewicz\inst{1}\\ Tomasz Bocheński\inst{1} and Przemysław Rokita\inst{1}}
%
\authorrunning{T. Trzciński, P. Andruszkiewicz, T. Bocheński and P. Rokita} 
%
\tocauthor{Tomasz Trzciński, Paweł Andruszkiewicz, Tomasz Bocheński and Przemysław Rokita`}
\institute{Warsaw University of Technology, Warsaw, Poland \\
\email{t.trzcinski@ii.pw.edu.pl, pandrus1@mion.elka.pw.edu.pl,\\tbochens@mion.elka.pw.edu.pl, pro@ii.pw.edu.pl} 
\and
Tooploox, Wrocław, Poland\\
}

\maketitle

\begin{abstract}
In this paper, we address the problem of popularity prediction of online videos shared in social media.  We prove that this challenging task can be approached using recently proposed deep neural network architectures. We cast the popularity prediction problem as a classification task and we aim to solve it using only visual cues extracted from videos. To that end, we propose a new method based on a Long-term Recurrent Convolutional Network (LRCN) that incorporates the sequentiality of the information in the model. Results obtained on a dataset of over 37'000 videos published on Facebook show that using our method leads to over 30\% improvement in prediction performance over the traditional shallow approaches and can provide valuable insights for content creators. 
\end{abstract}
%
\section{Introduction}
The problem of online content predicting popularity is extremely challenging since the popularity patterns are driven by many factors that are complex to model, such as social graph of the users, propagation patterns and content itself. The problem has therefore gained a lot of attention from the research community 
and various types of content were analyzed in this context, including but not limited to news articles~\cite{Bandari12,Castillo14}, Twitter messages~\cite{Hong11}, images~\cite{Khosla14} and videos~\cite{Pinto13,Tatar14,Chen16}.

In this paper, we focus on the problem of popularity prediction for videos distributed in social media. We postulate that the popularity of the video can be predicted with a higher accuracy if a prediction model incorporates the sequential character of the information presented in the video. To that end, we propose a method based on a deep recurrent neural network architecture called Long-term Recurrent Convolutional Network (LRCN)~\cite{Donahue15}, since in LRCN, the information about the sequentiality of the inputs is stored as parameters of the network. We are inspired by the successful application of this architecture to other video-related tasks, such as video activity recognition or video captioning. We call our method Popularity-LRCN and evaluate it on a dataset of over 37'000 videos collected from Facebook. Our results show that the Popularity-LRCN method leads to over 30\% improvement in prediction performance with respect to the state of the art.

The application of this method can be used by video content creators to optimize the chances that their work will become popular. In the context of everyday work, they often produce multiple videos on the same subject and choose the one that has the highest popularity potential based on their subjective opinion. Since our method predicts the popularity of the content based solely on the visual features, it can be applied on multiple videos before their publication to select the one that has the highest probability to become popular. 

The contributions of the paper are the following:
\begin{itemize}
\item an online video popularity prediction method that relies only on visual cues extracted from the video
\item application of a deep recurrent neural network architecture for the purpose of popularity prediction
\end{itemize}

The remainder of this paper is organized as follows: after a brief overview of the related work, we define the problem and describe the popularity prediction method we propose to address it. We then evaluate our method against the state of the art. 
Lastly, we conclude the paper with final word and future research.

\section{Related Work}

Popularity prediction of online content has gained a lot of attention within the research community due to the ubiquity of Internet and a stunning increase in the number of its users~\cite{Szabo10,Pinto13,Ramisa16}. 
None of the above mentioned works, however, considers using visual information for popularity prediction, while the method proposed in this paper relies solely on the visual cues.

The first attempt to use visual cues in the context of popularity prediction is the work of~\cite{Khosla14}, where they address the problem of popularity prediction for online images available on Flickr. Using a dataset of over 2 million images, the authors train a set of Support Vector Machines using visual features such as image color histogram or deep neural network outputs and apply them in the context of popularity prediction. Following this methodology, we use recently proposed neural network architectures for the purpose of video popularity prediction based on visual cues only.

The deep neural network architecture we use in this paper is known as Long-term Recurrent Convolutional Network (LRCN)~\cite{Donahue15} and is a combination of convolutional networks, initially used for image classification purposes, and Long-Short Term Memory (LSTM)~\cite{Hochreiter97} units, mostly used in the context of natural language processing and other text related tasks, where sequence of inputs plays an important role. Thanks to the combination of those two architectures LRCN was proven to be a good fit for several video-related tasks, such as activity recognition or video annotation. In this paper, we extend the list of potential applications of the LRCN-related architectures to the task of online video popularity prediction.

\section{Method}

In this section, we describe our method for popularity prediction of online videos that uses recurrent neural network architecture. We first define the popularity prediction problem as a classification task and then present the details of the deep neural network model that attempts to solve it.

\subsection{Problem}

We cast the problem of video popularity prediction as a binary classification task, similarly to~\cite{SimoSerra2015} where they analyze the popularity of fashion images using a classification approach. Our objective is to classify a video as popular or unpopular before its publication using only visual information from its frames.

More formally, let us a define a set of $N$ samples $\{x_i, l_i\}_{i=1}^{N}$, where $x_i$ is a video and $l_i \in \{0, 1\}$ is a corresponding popularity label equal to $1$, if a video is going to be popular after publication and $0$, if not. The label of the video is determined based on the number of times it has been watched by social network users after its publication, which we refer to as a {\it viewcount}. In other words, we aim to train a classifier $C$ that given a set of images assigns the sample with a predicted label $\hat{l} = f_C(x, \theta)$, where $\theta$ is a set of parameters of a classifier $C$. Our goal is to find the solution of the corresponding objective function:
\begin{equation}
\min_{\theta} \sum_{i=1}^{N} \left(\hat{l_i} - l_i \right)^2 = \min_{\theta} \sum_{i=1}^{N} \left(f_C(x, \theta) - l_i \right)^2
\end{equation}
which is a differentiable function that can be solved within the frames of deep neural network architectures, as we show next. 

\subsection{Popularity-LRCN} 

Since the problem definition presented in the previous section is quite general, a plethora of classifiers can be used to minimize it. In this section we propose an approach based on the recently proposed Long-term Recurrent Convolutional Network (LRCN) architecture~\cite{Donahue15}. We call this method Popularity-LRCN and show that its recurrent character, which incorporates the sequential character of the information displayed in the video frames, outperforms the methods that consider only individual frames. 




The architecture of the recurrent neural network used in Popularity-LRCN is a combination of a convolutional network with Long Short-Term Memory units, and is inspired by the architecture used in~\cite{Donahue15} for the task of activity recognition in videos. After initial experiments, we set the size of our network's input to 18 frames of size $227 \times 227 \times 3$ that represent the video. Our network consists of eight layers with learnable parameters. The first five layers are convolutional filters, followed by a fully-connected layer with 4096 neurons, LSTM and a final fully-connected layer with 2 neurons. 
We use soft-max as a classification layer. Since our method is trained to perform a binary classification, the output of the network is a 2-dimensional vector of probabilities linked to each of the popularity labels. The output is computed by propagating each of the video frames through the above architecture in a sequential manner and averaging the probability vector across all the frames, as done in~\cite{Donahue15}. 


To increase transitional invariance of our network, we use max pooling layers after the first, second and fifth convolutional layer. Local Response Normalization layers~\cite{Krizhevsky12} follow the first and the second max pooling layer. We use ReLU as a non-linear activation function and apply it to the output of every convolutional and fully-connected layer. To avoid overfitting, dropout layers are placed after the first fully-connected layer (with dropout ratio of 0.9) and after the LSTM layer (with dropout ratio of 0.5). 

When training, we input video frames of size $320 \times 240 \times 3$, crop it at random to $227 \times 227 \times 3$ and perform data augmentation with mirroring. We train our network for 12 epochs (30'000 iterations each) with batch size equal to 12. At prediction time, we also use data augmentation by cropping all frames from their four corners and around the center, as well as mirroring all of them. This way we generate 10 synthetical video representations from one test video representation. The final prediction output is the result of averaging probabilities across the frames and then across all generated representations.

\section{Evaluation}

In this section, we present the evaluation procedure used to analyze the performance of the proposed Popularity-LRCN method and compare it against the state of the art. We first present the dataset of over 37'000 videos published on Facebook, that is used in our experiments, and we describe how the popularity labels of the samples are assigned. We then discuss the evaluation criteria and baseline approaches. Lastly, we present the performance evaluation results. 

\subsection{Dataset}

Since our work is focused on predicting popularity of video content in social media, we collect the data from the social network with the highest number of users -- Facebook -- with reported 1.18 billion active everyday users worldwide\footnote{\url{http://newsroom.fb.com/company-info/}}.  To collect the dataset of the videos along with their corresponding viewcounts, we implemented a set of crawlers that use Facebook Graph API\footnote{\url{https://developers.facebook.com/docs/graph-api}} and ran it on the Facebook pages of over 160 video creators listed on a social media statistics website, TubularLabs\footnote{\url{http://tubularlabs.com/}}. To avoid crawling videos whose viewcounts are still changing, we restricted our crawlers to collect the data about videos that were online for at least 2 weeks. On top of the videos and their viewcounts, we also collected first and preferred thumbnails of the videos, as well as the number of publishers' followers. The Facebook videos missing any of this information were discarded. The resulting dataset consists of 37'042 videos published between June $1^{st}$ 2016 and September $31^{st}$ 2016.

\subsection{Labeling}

The distribution of the viewcounts across the videos from our dataset exhibits a large variation, as our dataset contains not only popular videos that were watched millions of times, but also those that were watched less than a few hundreds times. Similarly to~\cite{Khosla14}, we deal with this variation using log transform. Additionally, to reduce the bias linked to the fact that popular publishers receive more views on their videos irrespectably of their quality, we also include in our normalization procedure the number of followers of publishers' page. Our normalized popularity score is therefore computed according to the following formula:
\begin{equation}
	\mbox{normalized popularity score} = \log_2 \left( \frac{\mbox{viewcount} + 1}{\mbox{number of publisher's followers}}\right)
\end{equation}
The additional increment in the numerator prevents from computing logarithm of zero, in case of the videos that did not receive any views. 


After the normalization, we divide the dataset into two categories: popular videos and unpopular videos. To obtain an equal distribution of two classes in the training dataset, we split the dataset using median value of the normalized popularity score, following the approach of~\cite{SimoSerra2015}.

\subsection{Evaluation Criteria}

For the evaluation purposes, we used a $k$-fold train-test evaluation protocol with 5 folds. The Facebook dataset described above was split randomly and in each $k$-fold iteration 29'633 videos were used for training and 7'409 for testing. The performance metrics related to the classification accuracy and prediction quality were therefore averaged across 5 splits.

To evaluate the performances of the methods, we use classification accuracy, computed simply as a ratio of correctly classified samples on the test set. As a complementary performance metric we use Spearman correlation between video probabilities of belonging to the popular class and their normalized popularity score. This is because one of the goals of the proposed model is to help the creators to make an data-driven decision on which video should be published based on its predicted popularity. Since our model is trained to classify videos only as popular or unpopular, it provides a very granular evaluation. Therefore, we decided to use the probabilities of the popular class generated by the evaluated models as popularity scores and, following the evaluation protocol of~\cite{Khosla14} report their Spearman rank correlation with the normalized popularity score of the videos. We report the values of those metrics averaged across $k=5$ folds of our dataset along with the standard deviation values.

\subsection{Baselines}

Following the methodology presented in~\cite{Khosla14,Ramisa16}, we used as baselines traditional shallow classifiers, i.e. logistic regression classifier and SVM with RBF kernel. As input for these classifiers we used the following visual features that proven to be successful in other computer vision related tasks:

\begin{itemize}
\item HOG: a 8100-dimensional Histogram of Oriented Gradients descriptor~\cite{Dalal05}.

\item GIST: a 960-dimensional GIST descriptor, typically employed in image retrieval tasks~\cite{Douze09}

\item CaffeNet: we use as 1000-dimensional feature vectors activations of the last fully-connected layer of BVLC CaffeNet model\footnote{\scriptsize\url{https://github.com/BVLC/caffe/wiki/Model-Zoo}} (pre-trained on ILSVRC-2012\footnote{\scriptsize\url{http://www.image-net.org/challenges/LSVRC/}} object classification set).


\item ResNet: similarly to the above, we use 1000 activations of the last fully-connected layer of the ResNet-152~\cite{He15} neural network. 
\end{itemize}

\subsection{Results}

We evaluate the performances of all methods using the dataset of Facebook videos. Methods are implemented in Python using scikit-learn\footnote{\scriptsize\url{http://scikit-learn.org/}} and caffe\footnote{\scriptsize\url{http://caffe.berkeleyvision.org/}} libraries. It is worth noticing that Facebook increments a viewcount of a video after at least three seconds of watching it and by default turns the autoplay mode on. Therefore, in our experiments, we focus on the beginning of the video and take as input of the evaluated methods representative frames coming from the first six seconds of a video. In case of baseline features, we use early fusion approach and concatenate them before classification stage. For our Popularity-LRCN, we input raw RGB frames.


Tab.~\ref{tab:results} shows the results of the experiments. Popularity-LRCN clearly outperforms the competing shallow architectures, both in terms of classification accuracy and Spearman correlation by a large margin of 7\% and 34\%, respectively. We explain that by the fact that our Popularity-LRCN incorporates the sequentiality of the information, that is relevant in terms of popularity prediction, within the model. What is worth noticing, out of the competing methods, those using the activations of recently proposed convolutional neural networks perform better than those based on traditional features (such as GIST or HOG).



\begin{table}[t]
\setlength{\tabcolsep}{4pt}
\def\arraystretch{1}
\centering
\caption{Popularity prediction results. Classification accuracy is defined as a proportion of the videos correctly classified as popular or unpopular. Spearman correlation serves as a complementary quality evaluation metric and it is computed between the probability of a video belonging to a popular class and true normalized popularity score. We also report standard deviation values.}
\label{tab:results}
\begin{tabular}{cccc}
\toprule
Model & Features    &  Classification accuracy &  Spearman correlation\\ \bottomrule
& HOG & 0.587 $\pm$ 0.006 & 0.229 $\pm$ 0.014\\
logistic & GIST & 0.609 $\pm$ 0.007 & 0.321 $\pm$ 0.008\\
regression & CaffeNet & 0.622 $\pm$ 0.007 & 0.340 $\pm$ 0.007\\
 & ResNet & 0.645 $\pm$ 0.005 & 0.393 $\pm$ 0.010\\ \midrule
\multirow{4}{*}{SVM} & HOG & 0.616 $\pm$ 0.004 & 0.359 $\pm$ 0.008\\
 & GIST & 0.609 $\pm$ 0.006 & 0.294 $\pm$ 0.012 \\
 & CaffeNet & 0.653 $\pm$ 0.003 & 0.395 $\pm$ 0.007\\
 & ResNet & 0.650 $\pm$ 0.007 & 0.387 $\pm$ 0.015\\ \midrule 
 \multirow{1}{*}{Popularity-LRCN} & raw video frames & \multirow{1}{*}{\bf 0.7 $\pm$ 0.003} & \multirow{1}{*}{\bf 0.521 $\pm$ 0.009}\\ \bottomrule
 \end{tabular}
\end{table}

\begin{figure}[h!]
\centering
\subfloat[Best 100 videos]{\label{fig:best_100}\includegraphics[width=0.44\textwidth]{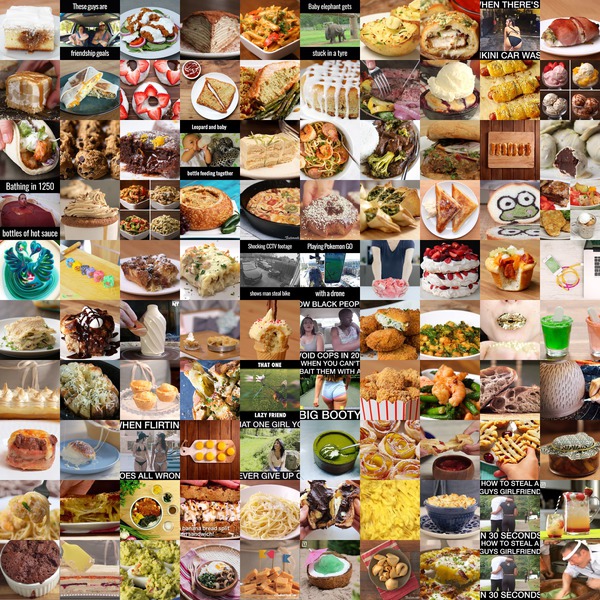}}\quad
\subfloat[worst 100 videos]{\label{fig:worst_100}\includegraphics[width=0.44\textwidth]{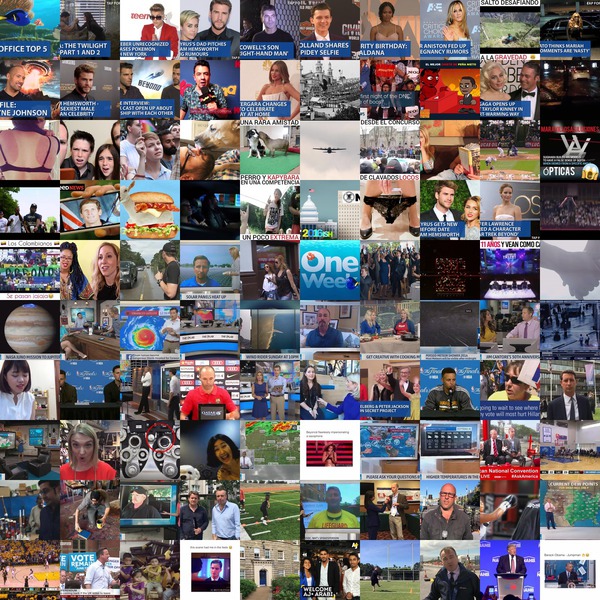}}
\caption{Results of Popularity-LRCN classification. A set of 100 thumbnails with the highest (a) and lowest (b) probability of popular class.}
\label{fig:best_worst}
\end{figure}

To visualize the results of the classification performed by Popularity-LRCN, in Fig.~\ref{fig:best_worst} we show 100 representative frames from the videos with the highest probability of belonging to the popular class and 100 frames from the videos with the lowest probability. As can be seen in Fig.~\ref{fig:best_100}, the videos whose frames contain pictures of food are classified as more probable to become popular, according to the Popularity-LRCN method. On the contrary, the lowest probability of becoming popular is linked to the videos with the interview-alike scenes and dull colors present in the opening scene. 

\section{Conclusions}

In this paper, we proposed a new approach to the task of online video content popularity prediction called Popularity-LRCN that relies on a deep recurrent neural network architecture. It uses only visual cues present in the representative frames of the video and outputs the predicted popularity class along with the corresponding probability. This method can therefore be used to compare videos in terms of their future popularity before their publication. To our best knowledge, this is the first attempt to address the video popularity prediction problem in this manner. 
Future research includes casting visual-based popularity prediction problem as a regression problem in the domain of video viewcounts and verifying the effectiveness of deep neural network architectures in this context.

\subsubsection*{Acknowledgment}
The authors would like to thank NowThisMedia Inc. for enabling this research with their hardware resources.

{

}

\end{document}